\documentclass[11pt]{article}
\usepackage[preprint]{acl}
\usepackage{times}
\usepackage{latexsym}
\usepackage[T1]{fontenc}
\usepackage[utf8]{inputenc}
\usepackage{microtype}
\usepackage{inconsolata}
\usepackage{graphicx}
\graphicspath{{figures/}}
\usepackage{amsmath,amssymb}
\usepackage{booktabs}

\title{Learning to Formalize: Bridging Open-Ended Workloads\\and Fast Decision Models}
\author{
  Chengxuan Pei\textsuperscript{1} \quad
  Ziyao Wang\textsuperscript{1} \quad
  Teng Mei\textsuperscript{2} \quad
  Zhe Li\textsuperscript{3} \\
  \textsuperscript{1}CUHK-Shenzhen \quad
  \textsuperscript{2}University of Tokyo \\
  \textsuperscript{3}Kunlunxin Corporation \\
  \small\texttt{chengxuanpei@link.cuhk.edu.cn} \quad
  \small\texttt{ziyaowang2@link.cuhk.edu.cn} \\
  \small\texttt{mei347957756@gmail.com} \quad
  \small\texttt{lizhe31@kunlunxin.com}
}

\begin{document}
\maketitle

\begin{abstract}
Decision models offer fast choices over supplied alternatives, while general-purpose large language models interpret open-ended tasks through flexible generation and reasoning at the cost of sequential decoding. Many workloads expect a generative backend to determine what can be done and therefore provide no well-defined candidate list. We bridge this interface gap with a small candidate formalizer trained on offline teacher-generated lists. The formalizer constructs explicit, validated alternatives from task context, and a frozen decision model selects an entry. A motivating comparison with official TypeSafe Jev-1.13.0 on 200 CALVIN-D test episodes reports 11\% success with generic motions versus 46.5\% with DeepSeek V4 Flash pose candidates. Separately, a six-arm privileged-state evaluation on 1,011 CALVIN-D validation annotations finds that a locally deployed decision model can achieve at most 52.6\% task success rate with the fine-tuned formalizer, compared with 11.8\% without it, and the end-to-end latency is kept to 2.09\,s, whereas Qwen3.5-27B is 3.67\,s with the exact experiment setting. 

\end{abstract}

\section{Introduction}

Many workloads delegate an open-ended instruction to a backend language model and expect it to determine the next action. A tool-using assistant must infer an operation and its arguments; a workflow system must turn a request into an actionable next step; a robot must translate a goal into a control target. In these settings, the workload supplies context and an execution interface, while useful alternatives remain implicit or loosely described. The backend is expected to analyze the situation, construct alternatives, and return a decision.

General-purpose large language models (LLMs) are attractive backends because they can interpret open-ended requests and support multi-step reasoning \citep{brown2020language,wei2022chain}. This flexibility has a latency cost: autoregressive decoding produces successive output tokens, and explicit intermediate reasoning can lengthen that sequence. Decision models offer a complementary interface. Their fast decision mode evaluates supplied alternatives and returns a typed choice without generating an extended answer \citep{almeida2026jev,autotrust2026jev27b}. However, this mode requires the application to specify what can be chosen. It does not automatically construct a missing list of executable alternatives, and benchmark evidence for Jev identifies weaknesses on some multi-step reasoning tasks \citep{li2026specialized}.

\begin{figure}[t]
\centering
\includegraphics[width=\columnwidth]{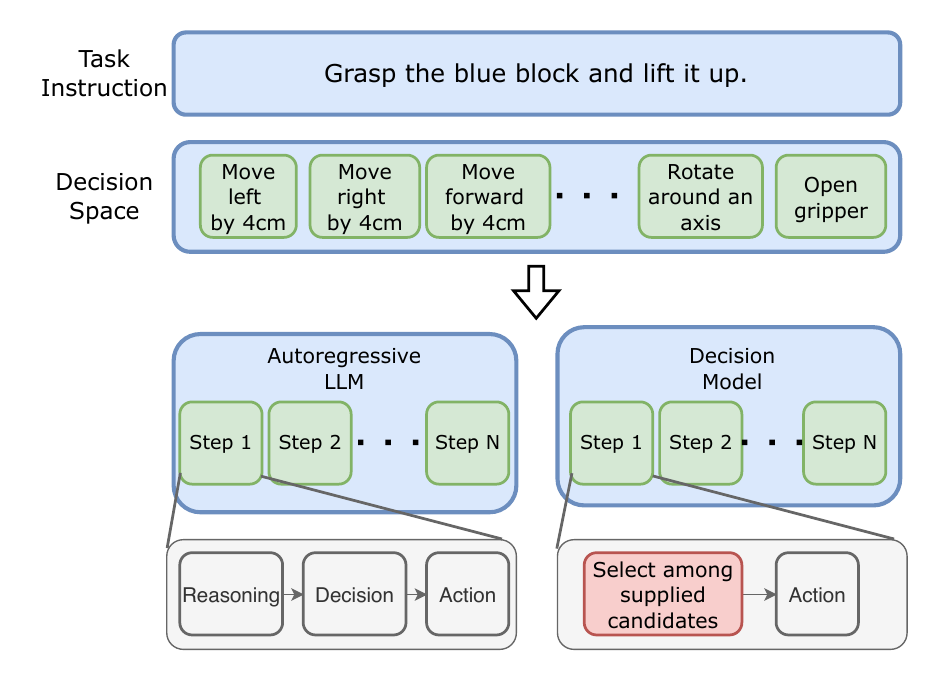}
\caption{The candidate-interface gap, illustrated with manipulation. The upper row sketches generic action primitives; task-conditioned candidates must instantiate useful next actions from the current state. The lower panels show a generative LLM workflow with explicit reasoning and a decision model selecting among supplied alternatives. Steps denote environment interactions; state observations are omitted for clarity. The distinction concerns the decision interface, without assuming that the selector lacks internal reasoning.}
\label{fig:decision-comparison}
\end{figure}

This creates an interface gap (Figure~\ref{fig:decision-comparison}). A workload built around an LLM may provide an instruction, state, and action schema without a ready-to-use choice list. The schema defines admissible operations, but a useful candidate set must instantiate task-relevant operations and parameters for the current state. Turning an action space into such a list requires semantic construction; listing generic primitives alone does not ensure useful choices. Even a strong selector cannot recover an action omitted from its candidate interface. The problem is therefore how to construct effective alternatives for a fast selector without paying for a large generative model at every request.

A matched test comparison makes this gap concrete (Section~\ref{sec:official-jev}). The native Jev baseline uses the official Jev-1.13.0 API to select among up to 34 generic motion choices. It succeeds on 22/200 CALVIN-D episodes (11\%). When DeepSeek V4 Flash generates three pose candidates for the same Jev selector, success rises to 93/200 (46.5\%), a gain of 35.5 percentage points. Both workflows use the same initial test episodes, observations, action budget, and success oracle. This supports the importance of task-conditioned candidate construction beyond exposing generic operations. 

We address this problem with a small learned \emph{candidate formalizer} followed by a frozen decision model. The formalizer interprets the workload context, organizes any available suggestions, and constructs a compact list of explicit alternatives in the application's schema. It supplies missing parameters when needed; a validator checks the resulting interface. The selector then compares these alternatives and returns an entry for the application to consume. This division assigns context interpretation and candidate construction to a small generative component, while retaining the short-output decision interface of a decision model.

The formalizer is trained through \emph{candidate-list distillation}. An offline teacher supplies structured candidate lists, and a smaller student learns to produce those lists from task context. The supervision captures the teacher's candidate-construction behavior. At deployment, the student need not generate a full reasoning transcript or make the final choice. Its task is narrower than solving the entire workload through unrestricted answer generation, although it still requires semantic interpretation rather than only JSON formatting. The intended benefit is useful decision quality with lower end-to-end latency than a large generative backend. Whether that benefit holds depends on both stages' costs and the quality of the exposed alternatives.

The interface is independent of a particular application: schemas can describe tool calls, workflow operations, control targets, or other structured choices. Existing options can be normalized when available; implicit options must be proposed when absent. Robotic manipulation is one example of the latter, rather than a definition of the method. Datasets, formalizer backbones, frozen selectors, and candidate counts are independent evaluation factors.

Our contributions are the candidate-interface gap as a deployment problem, a learned formalization stage trained on teacher candidate lists, and a six-arm evaluation of decision quality, latency, and resource use at two formalizer sizes. On all 1,011 CALVIN-D validation annotations, SFT Qwen3-4B and Qwen3-1.7B with frozen AutoTrust JEV-27B achieve 52.6\% and 52.1\% success, versus 11.8\% and 4.5\% for their base counterparts. The 1.7B pipeline takes 1.826\,s per decision, compared with 3.670\,s and 31.8\% success for direct Qwen3.5-27B generation: a 20.3-point success gain and 50.2\% lower mean decision latency. It also uses 14.2\% fewer reserved GPU-hours during evaluation. These results support the design in a privileged-state manipulation setting; broader datasets and controlled component ablations remain necessary to establish generality and attribute the gains.

\section{Background}
\label{sec:background}

\subsection{Open-Ended Workloads and Choice Interfaces}

We use \emph{decision model} to describe a component whose deployment role is to select among supplied alternatives. This role differs from generating an unrestricted answer or directly synthesizing a complete executable action. A fixed classification problem can enumerate its labels in advance. A parameterized action space, by contrast, may require constructing a small context-dependent subset before selection is practical. The action schema defines what can be expressed; the candidate list defines what the selector can consider at a particular decision. Figure~\ref{fig:decision-comparison} illustrates this distinction with manipulation primitives.

This distinction concerns the application interface, rather than a single model architecture. Autoregressive policies, value-based approaches, and explicit option selectors organize decision making differently. For example, Decision Transformer conditions sequence modeling on returns and trajectory history \citep{chen2021decision}, whereas a typed selector compares the options passed to it. The ability to rank those options does not by itself supply missing candidates or certify the outcome of executing one. An LLM-oriented workload can leave action construction to its backend, whereas a choice interface needs alternatives to be made explicit before selection. This mismatch motivates the formalization stage.

\subsection{Advantages of Decision Models}

Decision models provide an interface suited to repeated decisions inside software: a declared choice and, where supported, a distribution over admissible outcomes. These outputs can be consumed directly by application logic, reducing the need to extract a decision from unrestricted generated text. Jev implements this interface through typed probabilistic questions \citep{almeida2026jev}. The benefit is a constrained output contract; selecting a declared option still requires semantic correctness.

Low latency makes this interface attractive when decisions recur along an execution path. In an automated-gate setting, hosted Jev had a median request latency of 138\,ms, compared with 8.387\,s for Qwen3.6-27B using verbal generation and 0.997\,s using option likelihoods on an A100 \citep{rafe2026gates}. These are deployment-level measurements with different serving setups, so they illustrate achievable latency profiles rather than isolate an architectural speedup. Such results motivate avoiding unnecessary answer decoding for a well-scoped choice.

\subsection{Generative Reasoning, Decoding, and Distillation}

An autoregressive language model represents a structured response as a token sequence,
\begin{equation}
 p_{\theta}(y\mid x)=\prod_{j=1}^{|y|}p_{\theta}(y_j\mid x,y_{<j}).
\end{equation}
Language modeling supports flexible task specification and generation \citep{brown2020language}, but application software must interpret and validate the response. A list of tool-call arguments, workflow operations, or target poses can share a serialized interface while requiring different semantic checks. Syntactic validity alone does not establish that an action is applicable or useful. Generating intermediate steps can improve multi-step reasoning \citep{wei2022chain}, while increasing the number of sequential decoding steps. This is a motivation for specializing the online generative role, rather than assuming that all LLM requests have the same reasoning or latency requirements.

The Qwen3 family provides candidate student backbones \citep{yang2025qwen3}. Sequence-level knowledge distillation provides a precedent for supervising a student with teacher-produced sequences \citep{kim2016sequence}. Here, the target sequence contains competing alternatives for the same decision. They are neither consecutive steps in a plan nor several independent labels. The small model thereby learns to expose useful choices from open-ended context; the downstream selector remains responsible for choosing among them. A shorter output contract for the selector does not eliminate the cost of generating the list, which must be measured as part of the complete pipeline.

\subsection{Structured Decisions in Embodied Tasks}

Robotic manipulation provides a concrete setting in which an instruction leaves many intermediate actions unspecified. CALVIN supplies language-conditioned tasks and an accompanying manipulation dataset \citep{mees2022calvin}. Although its action schema is defined, it does not enumerate useful target poses for each encountered state. This illustrates a setting in which candidate construction precedes selection.

Related embodied approaches expose different interfaces. SayCan connects language preferences to executable skills and affordances \citep{ahn2022saycan}. Diffusion Policy models action generation through conditional denoising \citep{chi2023diffusion}, and OpenVLA predicts robot actions using pretrained vision-language representations \citep{kim2024openvla}. These approaches illustrate how observation formats and execution interfaces vary across applications. A structured-state adapter and an image-based policy expose different information, which must be accounted for when defining comparisons. Such workload-specific choices are independent of the candidate-list learning objective.

\section{Our Design}
\label{sec:design}

\subsection{Bridging the Candidate Interface}

The design separates two responsibilities that an open-ended LLM backend often combines: determining which concrete alternatives are worth considering, and deciding which alternative to use. A decision model supplies the second interface efficiently when an appropriate list is available. Our learned candidate formalizer supplies the first, using a small model adapted to the application's candidate schema.

Formalization includes semantic work. When a request contains informal suggestions, the model can make them comparable and supply explicit arguments. When the request supplies only a goal and state, it must construct plausible alternatives before serialization. For example, knowing that a tool accepts a file path does not specify which operation and path should be offered. The formalizer must expose relevant choices; the selector cannot repair an omitted alternative. We learn this narrower task from teacher candidate lists, keeping the final choice in a separate frozen component.

\subsection{Task Context and Formalized Candidates}

Consider an application $d$ with an allowed candidate schema and a mechanism for consuming a selected candidate. A candidate represents an actionable alternative, such as a parameterized tool call, a workflow operation, or a control target. The application supplies the schema; learning useful proposals does not require discovering the interface itself.

At decision step $t$, the context is
\begin{equation}
 x_t=(g,s_t,h_t,b_t),
\end{equation}
where $g$ specifies the goal, $s_t$ is the available task state, $h_t$ is observed interaction history, and $b_t$ contains any applicable budget information. A single-decision application may omit history and budget fields. Sequential applications update them after observing an outcome. Their precise contents are defined by the dataset adapter and documented for each experiment.

The small candidate formalizer, implemented as a generator $G_{\theta}$, constructs an ordered list,
\begin{equation}
 C_t=G_{\theta}(x_t)=[c_{t,1},\ldots,c_{t,K}].
\end{equation}
Available informal options can be included in $s_t$; the interface does not require an existing list. Each output entry contains an executable payload and may contain a short description of its intended purpose. All entries are alternatives for the same decision. Cardinality $K$ is configurable and held fixed within a controlled comparison. The current CALVIN adapter constructs candidates from state; formalizing pre-existing suggestions is a separate evaluation setting.

\subsection{Validation, Selection, and Application}

A validator $V_d$ checks a proposed list against the application's schema and declared constraints. Checks may include parseability, cardinality, required fields, argument types, bounds, and duplicate executable payloads. These checks are distinct from measuring semantic usefulness. For example, correctly typed arguments need not cause a tool to achieve the desired outcome, just as a bounded robot pose need not advance a manipulation task. An accompanying purpose description is an unverified model prediction.

For an accepted list, the frozen decision component selects
\begin{align}
 i_t &= \operatorname*{arg\,max}_{i\in\{1,\ldots,K\}}
 S_{\phi}(i\mid x_t,C_t),\\
 a_t &= \operatorname{payload}_{d}(c_{t,i_t}).
\end{align}
Here $S_{\phi}$ denotes the selector's preference rule; it need not expose calibrated task-success probabilities. The application consumes $a_t$ through its own execution or decision interface. Only the selected payload is consumed. In a sequential environment, the observed transition and feedback produce $x_{t+1}$ before a new candidate list is generated. In a single-decision setting, the process ends after the selected alternative is returned.

Validators and execution adapters define the application boundary. A new schema requires corresponding supervision and adapters; cross-domain transfer is not assumed. When a useful compact option list already exists, a learned formalizer may add unnecessary cost. The framework targets workloads where preparing that list requires interpretation.

Invalid generation and invalid selector indices require a documented recovery rule. Rejection, retries, or termination must be fixed for each task adapter before evaluation, with their effects included in the measured budget and latency. Silently excluding failed decisions would overstate pipeline reliability.

\begin{figure}[t]
\centering
\includegraphics[width=\columnwidth]{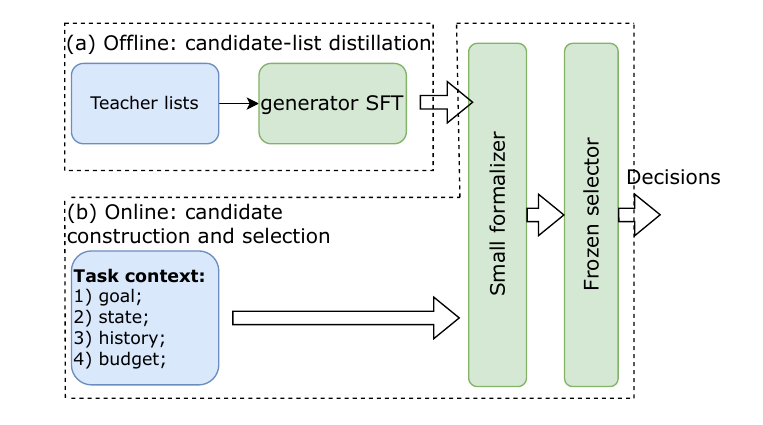}
\caption{Bridging an open-ended workload to a choice interface. Offline teacher candidate lists train the small formalizer through supervised fine-tuning. Online, the formalizer constructs explicit candidates from task context, and a frozen selector makes the decision. Application-specific validation, execution, and feedback are omitted for clarity.}
\label{fig:architecture}
\end{figure}

\subsection{Distilling Candidate Lists}

The offline teacher interprets task context and constructs alternatives in the target schema. This moves the recurring candidate-construction role from a large online backend to a specialized student. The recorded corpus is
\begin{equation}
 \mathcal{D}_{T}=\{(x^{(n)},y^{(n)})\}_{n=1}^{N},
\end{equation}
where $y^{(n)}$ serializes the teacher's candidate list for context $x^{(n)}$. The teacher is replaced at deployment, allowing expensive offline collection. Teacher identity, dataset, schema, and sampling settings define each collection configuration. Separate per-dataset generators and pooled training with schema conditioning are distinct experimental settings.

We train the generator using supervised next-token prediction on the teacher response:
\begin{equation}
\begin{split}
 \mathcal{L}_{\mathrm{SFT}}(\theta)=
 &-\frac{1}{N}\sum_{n=1}^{N}\sum_{j=1}^{|y^{(n)}|}\\
 &\log p_{\theta}(y_j^{(n)}\mid x^{(n)},y_{<j}^{(n)}).
\end{split}
\end{equation}
This expression treats the context as conditioning and supervises response tokens; the implementation must be checked for the corresponding loss mask. The target preserves the teacher's competing proposals rather than replacing the list with its preferred answer. It teaches the student what alternatives to expose without requiring a teacher reasoning trace or an online reasoning transcript. It does not establish that the student acquires the teacher's full reasoning ability.

The objective remains sequence supervision, rather than a permutation-invariant set loss. Candidate order and wording can therefore affect learning, and lower token loss need not imply better candidate coverage. Reproducing near-duplicate entries can be easier than preserving useful alternatives. Diversity, semantic validity, and downstream performance must consequently be measured separately. The method does not introduce a learned reward or an additional coverage objective.

\subsection{Independent Selection and the Latency Tradeoff}

For each experimental configuration, the chosen selector's parameters $\phi$ remain fixed throughout generator training. Different frozen selectors can be substituted through the same choice interface. Both $\theta$ and $\phi$ are frozen during evaluation. This separation permits controlled comparisons of candidate generators with the same selector, and comparisons of selectors on the same candidate lists. It also preserves a decision component independent of the generator, although independence of parameters does not guarantee that the selector will correct generation errors.

The efficiency hypothesis concerns total decision latency. For a formalizer emitting $m$ tokens, a simplified decomposition is
\begin{align}
 T_{\mathrm{form}} &= T_{\mathrm{prefill}}^{\mathrm{small}}
 + \sum_{j=1}^{m} T_{\mathrm{decode},j}^{\mathrm{small}},\\
 T_{\mathrm{total}} &= T_{\mathrm{form}}+T_{\mathrm{check}}+T_{\mathrm{select}}.
\end{align}
Here $T_{\mathrm{check}}$ includes validation and coordination; repairs add to the corresponding stage costs. Application execution is timed separately. The intended advantage occurs when $T_{\mathrm{total}}<T_{\mathrm{direct}}$, where $T_{\mathrm{direct}}$ is the latency of a large generative backend under the same workload. Specializing a smaller model can reduce the cost of candidate construction, while short-output selection avoids an extended answer-decoding stage. A list may still contain more tokens than a direct answer, so fewer output tokens are not assumed.

The selector must process the context and all alternatives. Increasing $K$, descriptions, or payload complexity can increase its input cost; a long formalizer response can also erase the intended savings. The experiment therefore measures both stages, complete decision latency, and task success together. Offline collection and training costs require a stated deployment workload for amortization. Lower latency alone does not establish lower total compute or monetary cost.

Failure can arise from omitted alternatives, poor selection, or unsuccessful execution. In sequential applications, outcomes can lead to contexts absent from teacher collection; observed feedback does not eliminate this distribution shift. Figure~\ref{fig:architecture} summarizes the training and deployment paths.

\section{Experiments}
\label{sec:experiments}

\subsection{Evaluation Scope and Research Questions}

Our evaluation asks whether candidate-list distillation improves decision quality and latency relative to base generators, how the tradeoff changes with formalizer size, and how the resulting pipelines compare with generic selection and direct LLM generation. We report two separate comparisons: an official TypeSafe Jev API study on the same 200 CALVIN-D test episodes, and a six-arm local comparison on all 1,011 official validation annotations in environment D. The first motivates task-conditioned candidates; the second evaluates learned formalization at two model sizes. CALVIN provides one instantiation of the general interface, rather than defining the method's scope.

The local comparison contains fine-tuned and base versions of Qwen3-4B and Qwen3-1.7B, each followed by the same frozen AutoTrust JEV-27B selector; JEV-27B selecting generic motion candidates; and Qwen3.5-27B directly producing actions. Each base/SFT pair controls the generator architecture and serving setup. The generic and direct arms compare deployment configurations with different candidate budgets or model families, rather than isolating a single component.

\subsection{Experimental Setup}
\label{sec:setup}

We evaluate all six arms on the same 1,011 CALVIN environment-D validation annotations across 34 manipulation tasks \citep{mees2022calvin}, excluded from SFT and checkpoint selection. Each annotation starts from fresh physics and is evaluated once per arm. The context contains the instruction, privileged state, recent actions and feedback, and remaining budget. Four episodes run concurrently with the official single-task success oracle, at most 12 decisions and 360 simulator steps, and 30-step action holds. One format repair is allowed; unrecovered failures terminate the episode and count against success. Physics pauses during inference.

DeepSeek supplies 36,453 training pairs from 4,612 episodes, each targeting a complete list of three pose candidates. Training and internal development use separate source recordings. We fully fine-tune Qwen3-4B and Qwen3-1.7B \citep{yang2025qwen3} for two epochs (4,558 updates), with effective batch size 16 and learning rate $10^{-5}$. Training uses FP32 parameters and optimizer state with BF16 autocast; inference uses unquantized BF16 weights. The AutoTrust JEV-27B selector \citep{autotrust2026jev27b} remains frozen. Hybrids supply three validated poses; generic selection uses up to 34 fixed poses through the publisher's wide-label extension. Direct Qwen3.5-27B \citep{qwen2026qwen35} generates one pose.

All four Qwen3 generators run sequentially on one A100 PCIe 40\,GB GPU with identical serving settings and matched initial prompts within each base/SFT pair. JEV-27B uses an A100 SXM4 80\,GB GPU and the publisher's selection protocol; direct Qwen3.5-27B subsequently uses that GPU. Requests use local endpoints. Startup, preflight, and warmup are excluded from evaluation timing.

\subsection{Motivating Comparison with the Official Jev API}
\label{sec:official-jev}

We first compare two candidate workflows using Jev-1.13.0 through the official TypeSafe API \citep{almeida2026jev} on the same 200 test episodes in CALVIN environment D \citep{mees2022calvin}. Both workflows use the same initial episodes, observation interface, action budget, and success oracle. The native Jev baseline selects from up to 34 generic motion choices and achieves 22/200 successes (11\%). When DeepSeek V4 Flash, accessed through its official API \citep{deepseek2026v4}, generates three pose candidates for Jev to select, the workflow achieves 93/200 successes (46.5\%; Table~\ref{tab:official-jev}). The observed difference is 35.5 percentage points, or approximately 4.23 times the success rate.

\begin{table}[t]
\centering
\small
\begin{tabular}{@{}p{0.52\columnwidth}rr@{}}
\toprule
Workflow & $K$ & Success\\
\midrule
Native Jev (generic motions) & $\leq34$ & 22/200 (11\%)\\
\textbf{DeepSeek poses $\rightarrow$ Jev} & 3 & \textbf{93/200 (46.5\%)}\\
\bottomrule
\end{tabular}
\caption{Official TypeSafe Jev-1.13.0 on the same 200 CALVIN-D test episodes, with matched observations, action budget, and success oracle. DeepSeek V4 Flash supplies three pose candidates through its official API. Candidate source and count both change. Latency is unreported. Bold highlights the workflow with higher observed success.}
\label{tab:official-jev}
\end{table}

This comparison provides empirical support for the candidate-interface gap in Figure~\ref{fig:decision-comparison}: exposing generic primitives from the action space does not yield the same workflow success as constructing task-conditioned pose alternatives. 



\begin{table*}[t]
\centering
\small
\begin{tabular}{@{}lrrrrrr@{}}
\toprule
Arm & $K$ & Success $\uparrow$ & Mean (s) & p50 (s) & p95 (s) & Output failures\\
\midrule
\textbf{SFT Qwen3-4B $\rightarrow$ JEV-27B} & 3 & \textbf{532/1011 (52.6\%)} & \textbf{2.573} & \textbf{2.543} & \textbf{3.066} & 15\\
Base Qwen3-4B $\rightarrow$ JEV-27B & 3 & 119/1011 (11.8\%) & 4.372 & 4.883 & 5.822 & 379\\
\textbf{SFT Qwen3-1.7B $\rightarrow$ JEV-27B} & 3 & \textbf{527/1011 (52.1\%)} & \textbf{1.826} & \textbf{1.838} & \textbf{2.276} & 16\\
Base Qwen3-1.7B $\rightarrow$ JEV-27B & 3 & 46/1011 (4.5\%) & 2.085 & 2.034 & 2.639 & 945\\
JEV-27B + generic motions & $\leq34$ & 296/1011 (29.3\%) & 5.602 & 5.737 & 5.803 & 0\\
Qwen3.5-27B direct & -- & 322/1011 (31.8\%) & 3.670 & 3.642 & 4.179 & 1\\
\bottomrule
\end{tabular}
\caption{Six-arm comparison on the same 1,011 CALVIN-D validation annotations, excluded from SFT and checkpoint selection. SFT denotes full-parameter candidate-list fine-tuning. Latency measures complete decisions, including parsing, validation, and repairs, and excludes application execution. Generic motions use a larger option set; direct generation returns one payload without a selector. Each annotation is evaluated once per arm. Bold highlights our pipelines and their success and latency improvements over direct generation.}
\label{tab:results}
\end{table*}

\subsection{Task Outcomes and Formalizer Size}

The SFT 4B and 1.7B pipelines complete 532 and 527 of 1,011 annotations (52.6\% and 52.1\%), compared with 119 (11.8\%) and 46 (4.5\%) for their base counterparts (Table~\ref{tab:results}). The corresponding gains, computed from counts, are 40.9 and 47.6 percentage points. Output failures decrease from 379 to 15 at 4B and from 945 to 16 at 1.7B. Candidate-list SFT therefore improves deployed pipeline behavior at both sizes, including interface reliability. The counts do not separate improvements in valid output, candidate usefulness, and selection.

Generic selection succeeds on 296 annotations (29.3\%) and direct Qwen3.5-27B on 322 (31.8\%). The SFT pipelines exceed direct generation by 20.8 percentage points at 4B and 20.3 at 1.7B. Their five-success difference is only 0.5 percentage points descriptively; we do not establish statistical equivalence. The smaller formalizer retains a similar observed success rate with a different latency and resource profile. Relative to generic candidates, both candidate content and count change; relative to direct generation, backbone, output target, and pipeline change. These comparisons do not isolate the selector's contribution.

\begin{table}[t]
\centering
\small
\begin{tabular}{@{}lrr@{}}
\toprule
Arm & First 200 & Remaining 811\\
\midrule
\textbf{SFT 4B hybrid} & \textbf{95 (47.5\%)} & \textbf{437 (53.9\%)}\\
Base 4B hybrid & 30 (15.0\%) & 89 (11.0\%)\\
\textbf{SFT 1.7B hybrid} & \textbf{96 (48.0\%)} & \textbf{431 (53.1\%)}\\
Base 1.7B hybrid & 12 (6.0\%) & 34 (4.2\%)\\
Generic + JEV & 70 (35.0\%) & 226 (27.9\%)\\
Direct 27B & 63 (31.5\%) & 259 (31.9\%)\\
\bottomrule
\end{tabular}
\caption{Success counts and rates on previously inspected annotations (first 200) and previously uninspected annotations (remaining 811). Both subsets belong to the official validation partition and are excluded from SFT and checkpoint selection. Bold highlights our pipelines.}
\label{tab:subsets}
\end{table}

On the previously uninspected 811 annotations, the SFT 4B and 1.7B pipelines achieve 437/811 (53.9\%) and 431/811 (53.1\%), compared with 259/811 (31.9\%) for direct generation (Table~\ref{tab:subsets}). The descriptive advantage is present beyond the first 200 inspected annotations; both subsets are excluded from training and checkpoint selection. Per-task outcomes are heterogeneous: the SFT pipelines complete 23/29 and 24/29 stacking annotations, versus 3/29 for direct generation, but neither succeeds on any of the 29 drawer-closing annotations, where base 4B succeeds on 14. Aggregate success does not imply uniform improvement across tasks.

\subsection{Latency and Request-Level Analysis}

Mean decision latency is 2.573\,s for the SFT 4B pipeline and 1.826\,s for SFT 1.7B, compared with 3.670\,s for direct generation: reductions of 29.9\% and 50.2\%, respectively. Their p95 latencies are 3.066\,s and 2.276\,s, versus 4.179\,s for direct generation. Relative to their base generators at 4.372\,s and 2.085\,s, SFT reduces mean decision latency by 41.1\% and 12.4\%. Decision latency includes all requests, parsing, validation, and repairs, and excludes application execution.

\begin{table*}[t]
\centering
\small
\begin{tabular}{@{}llrrrrrr@{}}
\toprule
Generator & Stage & Requests & Mean (s) & p95 (s) & Prefill (s) & Decode (s) & Out tok.\\
\midrule
\textbf{SFT 4B} & Generate & 7948 & \textbf{1.844} & \textbf{2.023} & 0.043 & \textbf{1.786} & \textbf{201.8}\\
\textbf{SFT 4B} & Select & 7916 & 0.725 & 1.201 & 0.644 & 0.000 & 1.0\\
Base 4B & Generate & 13628 & 2.365 & 2.590 & 0.036 & 2.311 & 259.9\\
Base 4B & Select & 8189 & 0.634 & 1.098 & 0.576 & 0.000 & 1.0\\
\textbf{SFT 1.7B} & Generate & 7949 & \textbf{0.937} & \textbf{1.030} & 0.020 & \textbf{0.902} & \textbf{201.7}\\
\textbf{SFT 1.7B} & Select & 7915 & 0.887 & 1.336 & 0.749 & 0.000 & 1.0\\
Base 1.7B & Generate & 3242 & 1.221 & 1.325 & 0.017 & 1.189 & 256.2\\
Base 1.7B & Select & 1320 & 0.574 & 0.988 & 0.522 & 0.000 & 1.0\\
\bottomrule
\end{tabular}
\caption{Request-level timing of the Qwen3 formalizers with frozen JEV-27B selection. Mean and p95 are request latencies; prefill and decode are mean server inference components. Repairs add generator requests, while failed generation skips selection, so stage means cannot be added to recover mean decision latency. Out tok. denotes mean output tokens. Bold generation values improve over the corresponding base model.}
\label{tab:timing}
\end{table*}

Request-level statistics in Table~\ref{tab:timing} show SFT generation means of 1.844\,s at 4B and 0.937\,s at 1.7B; selection means are 0.725\,s and 0.887\,s. Thus both online stages contribute materially to latency. Generation remains dominated by decoding: mean server decode time is 1.786\,s at 4B and 0.902\,s at 1.7B. The selector reports one output token and no decode time, but still incurs input processing, queueing, and transport. Selection time varies despite a fixed checkpoint because requests and concurrent trajectories differ.

Request and decision means have different denominators. Repairs add generation requests, whereas failed generation skips selection; adding stage means would therefore misestimate complete decision latency. Base 4B issues 13,628 generation requests versus 7,948 for SFT 4B. Base 1.7B issues only 3,242 because failures terminate many episodes early, compared with 7,949 for SFT 1.7B. Its mean episode wall time of 4.814\,s is consequently much shorter than SFT 1.7B's 15.115\,s despite much lower success. Episode wall time alone is not a task-efficiency measure.

The initial generator requests provide a closer match: all 1,011 paired cases have identical prompts and initial states within each size. At 4B, both arms average 820.0 input tokens; SFT takes 1.853\,s and emits 204.5 tokens, versus 2.281\,s and 252.1 tokens for base. At 1.7B, both average 824.0 input tokens; SFT takes 0.948\,s and emits 204.4 tokens, versus 1.233\,s and 260.0 tokens for base. The respective request-time reductions are 18.8\% and 23.1\%. This removes prompt and state differences but retains differences in generated length and concurrent batching; it is not an isolated throughput microbenchmark.

The generic baseline has mean decision latency 5.602\,s and mean selector input length 4,567.5 tokens, versus approximately 1,427 tokens for SFT hybrids. Its mean server queue time is 2.327\,s under four concurrent episodes. These differences prevent attributing its latency to candidate construction alone. Direct generation averages 3.633\,s per request and 3.670\,s per complete decision, with 3.155\,s of server decode time per request. Server inference excludes HTTP transport and is not pure CUDA kernel time; measurements depend on hardware and concurrency.

\subsection{Resource Use}

\begin{table}[t]
\centering
\small
\begin{tabular}{@{}lrrr@{}}
\toprule
Arm & Min. & GPU-h & kWh\\
\midrule
\textbf{SFT 4B hybrid} & \textbf{92.64} & 3.0879 & 0.6627\\
Base 4B hybrid & 163.56 & 5.4519 & 1.0298\\
\textbf{SFT 1.7B hybrid} & \textbf{67.12} & \textbf{2.2375} & \textbf{0.5182}\\
Base 1.7B hybrid & 22.32 & 0.7439 & 0.1389\\
Generic + JEV & 255.39 & 4.2565 & 1.0225\\
Direct 27B & 156.40 & 2.6066 & 0.6244\\
\bottomrule
\end{tabular}
\caption{Evaluation elapsed time, reserved GPU-hours, and whole-board NVML energy. Hybrids reserve two boards, including waiting; other arms reserve one. Energy excludes CPU and other system components. Four episodes run concurrently. Early output failures can shorten evaluation. Bold values indicate improvements over direct generation.}
\label{tab:resources}
\end{table}

Table~\ref{tab:resources} reports evaluation elapsed time, reserved GPU-hours, and whole-board NVML energy. SFT 1.7B uses 2.2375 GPU-hours and 0.5182\,kWh, compared with 2.6066 GPU-hours and 0.6244\,kWh for direct generation: reductions of 14.2\% and 17.0\%. SFT 4B uses 3.0879 GPU-hours and 0.6627\,kWh, respectively 18.5\% and 6.1\% more than direct generation. Thus the lower decision latency does not imply uniformly lower resource use; the smaller formalizer provides the more favorable measured tradeoff in this run.

All hybrids reserve two boards, including waiting time; generic selection and direct generation reserve one. Base 1.7B uses only 0.7439 GPU-hours and 0.1389\,kWh, but its 945 output failures end most episodes early. This low evaluation cost accompanies only 46 successes and does not establish an efficient successful policy.

Local inference incurs no API fee, but hardware, system energy, teacher collection, and SFT costs remain. A monetary or amortized comparison requires hardware rates, training costs, and a deployment workload. The measured latency advantage does not establish universal compute savings.

\subsection{Discussion}

The completed six-arm study supports candidate-list adaptation at two formalizer sizes under a fixed selector and protocol. It does not establish transfer across datasets or selector families. Additional datasets should document their schemas, splits, metrics, and adaptation separately. Repeated runs and dependence-aware uncertainty are needed to assess stability across overlapping annotations.

Further controls should match candidate count, observations, and failure handling. Fixed-list selector comparisons, random or first-candidate selection, matched-protocol teacher proposals, and matched-backbone direct generation would distinguish component contributions. Candidate order, descriptions, history, and budget fields also remain unablated. For pre-existing informal alternatives, deterministic schema conversion would separate normalization from semantic construction. Candidate-count and output-length sweeps can test the tradeoff as lists become more complex.

\section{Conclusion}

Fast decision models require explicit choices, whereas open-ended workloads often delegate candidate construction to a generative LLM. We bridge this gap with a small candidate formalizer trained on teacher-generated lists and a frozen selector that makes the final decision. The official Jev-1.13.0 comparison motivates task-conditioned candidate construction. Separately, the completed six-arm CALVIN-D evaluation shows gains from candidate-list SFT at both 4B and 1.7B sizes. The 1.7B pipeline achieves 52.1\% success at 1.826\,s per decision, versus 31.8\% at 3.670\,s for direct Qwen3.5-27B generation, with lower measured evaluation GPU-hours and board energy. The 4B pipeline achieves 52.6\% success but uses more resources than direct generation, illustrating the importance of formalizer size. These results support the success--latency tradeoff on validation episodes excluded from training and checkpoint selection; broader datasets and controlled component comparisons are needed to establish its wider benefit.

\section*{Limitations}

Our evaluation covers CALVIN-D manipulation with privileged state and paused physics, so it does not establish real-time performance or transfer across datasets, schemas, and selector families. Normalizing pre-existing informal alternatives remains unevaluated. Each annotation is evaluated once per arm, and annotations may overlap; repeated runs and dependence-aware uncertainty are needed. Candidate lists can omit useful actions or inherit teacher errors, and structural validation does not certify successful execution.

Generic and generated candidates differ in count, while direct generation uses a different model family and output target. These comparisons establish pipeline-level tradeoffs without isolating candidate quality, output validity, or selector contribution. The official-API study provides no latency measurements. Resource savings depend on formalizer size: SFT 1.7B uses fewer evaluation GPU-hours and less board energy than direct generation, whereas SFT 4B uses more. Costs exclude teacher collection, SFT, and CPU/system energy, and early output failures can shorten evaluation runs.

\bibliography{reference}

@article{ahn2022saycan,
  title={Do As I Can, Not As I Say: Grounding Language in Robotic Affordances},
  author={Ahn, Michael and Brohan, Anthony and Brown, Noah and Chebotar, Yevgen and Cortes, Omar and David, Byron and Finn, Chelsea and Fu, Chuyuan and others},
  journal={arXiv preprint arXiv:2204.01691},
  year={2022},
  url={https://arxiv.org/abs/2204.01691}
}

@article{mees2022calvin,
  title={{CALVIN}: A Benchmark for Language-Conditioned Policy Learning for Long-Horizon Robot Manipulation Tasks},
  author={Mees, Oier and Hermann, Lukas and Rosete-Beas, Erick and Burgard, Wolfram},
  journal={IEEE Robotics and Automation Letters},
  volume={7},
  number={3},
  pages={7327--7334},
  year={2022},
  url={https://arxiv.org/abs/2112.03227}
}

@article{chen2021decision,
  title={Decision Transformer: Reinforcement Learning via Sequence Modeling},
  author={Chen, Lili and Lu, Kevin and Rajeswaran, Aravind and Lee, Kimin and Grover, Aditya and Laskin, Michael and Abbeel, Pieter and Srinivas, Aravind and Mordatch, Igor},
  journal={Advances in Neural Information Processing Systems},
  volume={34},
  year={2021},
  url={https://arxiv.org/abs/2106.01345}
}

@article{chi2023diffusion,
  title={Diffusion Policy: Visuomotor Policy Learning via Action Diffusion},
  author={Chi, Cheng and Xu, Zhenjia and Feng, Siyuan and Cousineau, Eric and Du, Yilun and Burchfiel, Benjamin and Tedrake, Russ and Song, Shuran},
  journal={arXiv preprint arXiv:2303.04137},
  year={2023},
  url={https://arxiv.org/abs/2303.04137}
}

@article{brown2020language,
  title={Language Models are Few-Shot Learners},
  author={Brown, Tom B. and Mann, Benjamin and Ryder, Nick and Subbiah, Melanie and Kaplan, Jared and Dhariwal, Prafulla and Neelakantan, Arvind and Shyam, Pranav and others},
  journal={Advances in Neural Information Processing Systems},
  volume={33},
  year={2020},
  url={https://arxiv.org/abs/2005.14165}
}

@article{kim2024openvla,
  title={{OpenVLA}: An Open-Source Vision-Language-Action Model},
  author={Kim, Moo Jin and Pertsch, Karl and Karamcheti, Siddharth and Xiao, Ted and Balakrishna, Ashwin and Nair, Suraj and Rafailov, Rafael and Foster, Ethan and Lam, Grace and Sanketi, Pannag and Vuong, Quan and Kollar, Thomas and Burchfiel, Benjamin and Tedrake, Russ and Sadigh, Dorsa and Levine, Sergey and Liang, Percy and Finn, Chelsea},
  journal={arXiv preprint arXiv:2406.09246},
  year={2024},
  url={https://arxiv.org/abs/2406.09246}
}

@article{yang2025qwen3,
  title={{Qwen3} Technical Report},
  author={Yang, An and Li, Anfeng and Yang, Baosong and Zhang, Beichen and Hui, Binyuan and Zheng, Bo and Yu, Bowen and Gao, Chang and others},
  journal={arXiv preprint arXiv:2505.09388},
  year={2025},
  url={https://arxiv.org/abs/2505.09388}
}

@inproceedings{kim2016sequence,
  title={Sequence-Level Knowledge Distillation},
  author={Kim, Yoon and Rush, Alexander M.},
  booktitle={Proceedings of the 2016 Conference on Empirical Methods in Natural Language Processing},
  year={2016},
  pages={1317--1327},
  publisher={Association for Computational Linguistics},
  address={Austin, Texas},
  doi={10.18653/v1/D16-1139},
  url={https://aclanthology.org/D16-1139/}
}

@misc{almeida2026jev,
  title={Introducing System One Models \& {Jev}},
  author={Almeida, Diogo},
  year={2026},
  month=sep,
  howpublished={TypeSafe AI developer blog},
  url={https://typesafe.ai/blog/introducing-system-one-models-and-jev},
  note={Published September 15, 2026. Accessed October 24, 2026}
}

@article{li2026specialized,
  title={Specialized Decision Models vs. General-Purpose {LLMs}: Benchmarking {Jev} Across Knowledge, Reasoning, and Multilingual Tasks},
  author={Li, Xing and Chang, Qingcheng and Ning, Jinzhong and Xu, Changfeng and Zhang, Shenlong and Zhang, Yijia and Luo, Ling and Lin, Hongfei},
  journal={arXiv preprint arXiv:2610.11978},
  year={2026},
  url={https://arxiv.org/abs/2610.11978v1}
}

@article{rafe2026gates,
  title={Benchmarking System One decision models against trained classifiers and language models for automated decision gates},
  author={Rafe, Amir and Das, Subasish},
  journal={arXiv preprint arXiv:2610.00346},
  year={2026},
  url={https://arxiv.org/abs/2610.00346v2}
}

@misc{autotrust2026jev27b,
  title={{JEV-27B}: Model Card},
  author={{AutoTrust AI Lab}},
  year={2026},
  howpublished={Hugging Face model repository},
  url={https://huggingface.co/autotrust/JEV-27B},
  note={Accessed October 24, 2026}
}

@misc{qwen2026qwen35,
  title={{Qwen3.5-27B}: Model Card},
  author={{Qwen Team}},
  year={2026},
  howpublished={Hugging Face model repository},
  url={https://huggingface.co/Qwen/Qwen3.5-27B},
  note={Accessed October 24, 2026}
}

@article{wei2022chain,
  title={Chain-of-Thought Prompting Elicits Reasoning in Large Language Models},
  author={Wei, Jason and Wang, Xuezhi and Schuurmans, Dale and Bosma, Maarten and Ichter, Brian and Xia, Fei and Chi, Ed and Le, Quoc and Zhou, Denny},
  journal={Advances in Neural Information Processing Systems},
  volume={35},
  year={2022},
  url={https://arxiv.org/abs/2201.11903}
}

@misc{deepseek2026v4,
  title={{DeepSeek-V4} Preview Release},
  author={{DeepSeek}},
  year={2026},
  howpublished={Official DeepSeek API documentation},
  url={https://api-docs.deepseek.com/news/news260424/},
  note={Describes the V4 Pro and V4 Flash models. Accessed October 24, 2026}
}
\end{document}